\documentclass{svproc}
\usepackage{url}

\usepackage{graphicx}
\usepackage{hyperref}
\usepackage{adjustbox}
\usepackage[table]{xcolor}
\usepackage{times}
\usepackage{latexsym}
\usepackage{rotating}

\usepackage[T1]{fontenc}

\usepackage[utf8]{inputenc}

\usepackage{microtype}

\usepackage{inconsolata}
\usepackage{float}
\usepackage{comment}
\begin{document}
\mainmatter              

\title{Improving Cooperation in  Collaborative Embodied AI }

\authorrunning{Hima Jacob et al.} 
\tocauthor{Hima Jacob Leven Suprabha, 
 Laxmi Nag Laxminarayan Nagesh,
Ajith Nair,
 Alvin Reuben Amal Selvaster, 
 Ayan Khan,
 Raghuram Damarla,
 Sanju Hannah Samuel,
 Sreenithi Saravana Perumal,
 Titouan Puech,
 Venkataramireddy Marella,
Vishal Sonar,  Alessandro Suglia, and Oliver Lemon} 

  \author{Hima Jacob Leven Suprabha \and 
 Laxmi Nag Laxminarayan Nagesh \and
Ajith Nair \and  
 Alvin Reuben Amal Selvaster \and 
 Ayan Khan \and
 Raghuram Damarla \and
 Sanju Hannah Samuel \and
 Sreenithi Saravana Perumal \and
 Titouan Puech \and
 Venkataramireddy Marella \and
Vishal Sonar \and  Alessandro Suglia \and Oliver Lemon
}

\institute{School of Mathematical and Computer Sciences,\\ Heriot-Watt University, Edinburgh, UK,\\
\email{o.lemon@hw.ac.uk},\\ 
\texttt{http://www.macs.hw.ac.uk/InteractionLab}
}

\maketitle
\begin{abstract}
The integration of Large Language Models (LLMs) into multiagent systems has opened new possibilities for collaborative reasoning and cooperation with AI agents.
This paper explores different prompting methods and  evaluates their effectiveness in enhancing agent
collaborative behaviour and decision-making. We enhance CoELA, a framework designed for building Collaborative Embodied Agents that leverage LLMs for multi-agent communication, reasoning, and task coordination in shared virtual spaces. Through systematic experimentation, we examine different  LLMs and prompt engineering strategies to identify optimised combinations that maximise collaboration performance. Furthermore, we extend our research by integrating speech capabilities, enabling seamless collaborative voice-based interactions. Our findings highlight the effectiveness of prompt optimisation in enhancing collaborative agent performance; for example, our best combination improved the  efficiency of the system running with Gemma3 by 22\% compared to the original CoELA system. In addition, the speech integration provides a more engaging user interface for iterative system development and demonstrations.\\
\href{https://youtu.be/MtP8BH7kYsE}{Alice and Bob – Voice Chat GUI Video demonstration.}

\href{https://github.com/Himajacob/Co-LLM-Agents}{Github link to codebase.}

\keywords{Collaborative Embodied AI, co-operative AI, LLMs, evaluation}

\end{abstract}

\section{Introduction}

As artificial intelligence continues to evolve, there is growing interest in enabling agents not just to act intelligently on their own but to work together -- collaboratively solving problems, sharing tasks, and communicating effectively. This vision is central to Collaborative Embodied AI (CEAI), where virtual or physical agents operate in shared environments, coordinating their actions in pursuit of common goals \cite{deitke2022retrospectives,umass2023coela,zhang2024buildingcooperativeembodiedagents,mandi2023rocodialecticmultirobotcollaboration,wang2023voyageropenendedembodiedagent}. 

With the rise of Large Language Models (LLMs), a new set of possibilities has emerged. LLMs are capable of understanding nuanced instructions, engaging in conversation, and even generating plans -- making them ideal candidates for powering collaborative agents. However, integrating LLMs into embodied agents presents unique challenges: how can these models reason about the environment? How should agents communicate effectively? And what types of prompts or memory structures are most useful in team-based scenarios? 

\begin{figure*}[htb]
    \centering
    \includegraphics[width=0.99\textwidth]{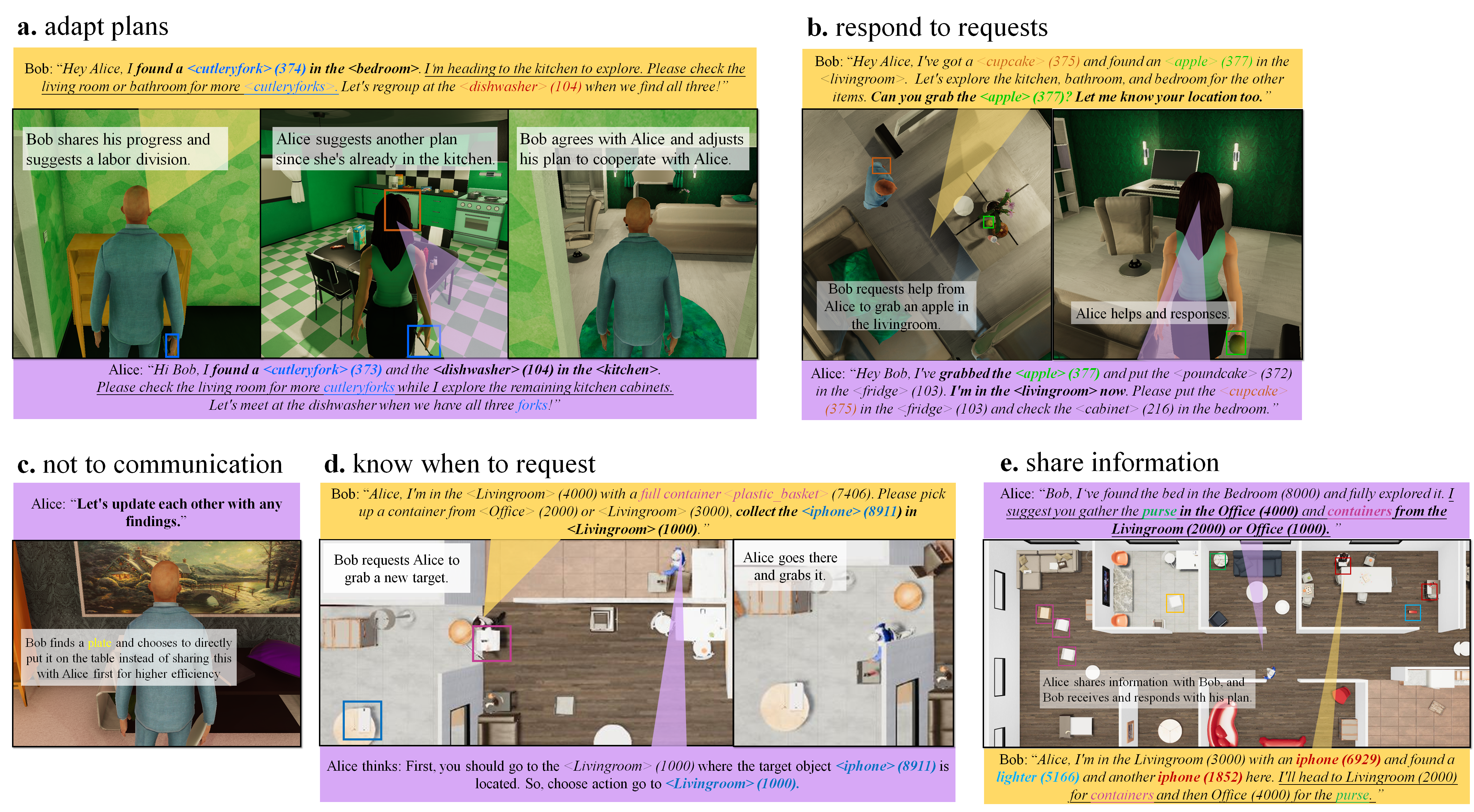}
    \caption{Agents collaborating in a simulation environment  (CoELA) \cite{umass2023coela} }
    \label{fig:coela-examples}
\end{figure*}

This work builds on the CoELA \cite{umass2023coela,zhang2024buildingcooperativeembodiedagents} framework -- a modular system for creating multi-agent embodied AI -- exploring how different prompting strategies and LLMs affect coordination, decision-making, and task performance.
See figure \ref{fig:coela-examples} for examples of agents collaborating in CoELA.

Building  on CoELA, we develop new structured and communication-focused prompts and evaluate them with models such as  Llama 3.1 \cite{grattafiori2024llama3herdmodels}, DeepSeek r1 \cite{deepseekai2025deepseekr1incentivizingreasoningcapability}, Mistral \cite{jiang2023mistral7b}, and Gemma3 \cite{gemmateam2025gemma3technicalreport}.
By experimenting across multiple configurations, we aim to uncover the best combinations of LLMs and prompts for efficient teamwork. Ultimately, this work contributes to the broader goal of building embodied agents that interact not just intelligently, but collaboratively, much like humans do in real-world tasks.

\section{Background and Related Work}

The field of Collaborative Embodied AI (CEAI) focuses on building agents that not only perceive and act within their environments but also cooperate and communicate with other agents -- be they human or artificial. The convergence of Embodied AI and LLMs has opened a promising path for designing such agents \cite{deitke2022retrospectives}.

\subsection{Foundations of Collaborative Embodied AI}
Collaborative Embodied AI (CEAI) functions as an interdisciplinary research area combining elements of robotics with embodied reasoning and multi-agent collaboration techniques. CEAI agents differ from conventional AI systems because they are built to dynamically interact and collaborate with humans and artificial agents while performing perception and action tasks.

The system CoELA (``Collaborative Embodied Language Agents'') \cite{umass2023coela,zhang2024buildingcooperativeembodiedagents}  represents a primary framework in the field by providing a modular cognitive architecture that enables decentralised agent cooperation within three-dimensional spaces. The decision-making capabilities of the system emerge from structured interactions among five essential components: Perception, Memory, Planning, Communication, and Execution. CoELA maintains a  balance between high-level task planning and low-level action execution, which aims to support operational efficiency despite facing costly communication constraints.

\subsection{Language as a Tool for Reasoning and Coordination}
CEAI has experienced substantial advances through the use of Large Language Models (LLMs), which function as cognitive engines to augment agents' capabilities in planning, acting, and communicating. LLM-driven agents merge reasoning with action, which surpasses traditional rule-based control systems by enabling intelligent and adaptable decision-making. Techniques like ReAct \cite{yao2023reactsynergizingreasoningacting}   show how agents can think out loud and decide actions in a dynamic way, rather than relying on static plans, while Reflexion \cite{shinn2023reflexionlanguageagentsverbal}   adds self-feedback loops, where an agent can reflect on its past performance and adjust future behaviour.

Chain-of-thought prompting and self-consistency strategies have improved clarity and robustness in multi-step reasoning. The CAMEL framework \cite{li2023camelcommunicativeagentsmind} introduced role-playing dialogues for coordination, and the Voyager system \cite{wang2023voyageropenendedembodiedagent} enabled agents to learn continuously in open-world settings by refining code and tasks dynamically. For planning tasks, structured approaches such as DESP \cite{wang2024describeexplainplanselect} and Tree of Thoughts \cite{hassouna2024llm} treat planing as a multistage or branching process which improves the efficiency of the decision-making process.

Recent frameworks such as Agent-S \cite{agashe2024agents} have further demonstrated how multimodal LLM agents can autonomously acquire task knowledge and refine their strategies via direct interaction with GUI environments, without relying on human supervision or explicit feedback.

\subsection{Memory, Simulation, and Human-AI Interaction}
A key obstacle in CEAI involves maintaining context throughout extended multi-turn interactions.  LLM-powered agents approach  this issue by using sophisticated memory structures. According to \cite{zhang2024buildingcooperativeembodiedagents}, CoELA operates with a memory system organised into three distinct levels:

\begin{itemize}
    \item Semantic memory (storing factual world knowledge)

    \item Episodic memory (allows tracking of previous interactions and actions)

    \item Procedural memory (functions as the repository of learnt behaviours together with execution strategies.)
\end{itemize}
The model draws upon  cognitive models such as Soar and ACT-R to enhance the long-term adaptive capabilities of LLM-powered agents. The LLM-Agent-UMF framework developed in \cite{hassouna2024llm} combines both active and passive memory models within modular memory systems to improve decision-making processes.

LLM-powered dialogue systems enable clearer and faster interactions between multiple agents in human-agent collaboration settings. The systems AutoGen \cite{autogen2023} and ROSA \cite{royce2025enablingnovelmissionoperations} concentrate on dialogue-based natural interactions which improve both communication clarity and adaptability. Research on CoELA demonstrates that using natural language communication boosts human trust and task efficiency which leads to a preference for LLM agents over conventional symbolic or template-based systems. The enhancements show significant benefits in complex simulated settings like VirtualHome \cite{puig2018virtualhome} and Habitat 3.0 \cite{puig2023habitat30cohabitathumans}, which contain multi-room task-orientated scenarios that serve as realistic research platforms for CEAI studies.

\subsection{Comparison with Prior Work}

The embodied AI systems Voyager \cite{wang2023voyageropenendedembodiedagent} and Reflexion \cite{shinn2023reflexionlanguageagentsverbal} demonstrate their capabilities using single agents within open-ended settings, whereas our research highlights multi-agent teamwork with role-specific negotiation.

CAMEL \cite{li2023camelcommunicativeagentsmind} uses role-playing techniques to simulate how large language models interact socially, but does not address  execution problems in physical environments.

The ``Tree of Thoughts" model \cite{yao2023treethoughtsdeliberateproblem} introduces a formal reasoning framework yet does not to connect this with practical application by physical agents in tangible tasks.

 The RoCo system \cite{mandi2023rocodialecticmultirobotcollaboration} uses LLM-guided dialogue to facilitate multi-robot collaboration but does not incorporate integrated memory and semantic perception which results in agents that do not adequately manage temporal constraints.

AutoGen from Microsoft \cite{autogen2023}, alongside ROSA \cite{royce2025enablingnovelmissionoperations}, investigates interpretability and human-AI communication alignment in multi-agent systems, which parallels our structured prompting research for reliable collaboration. While frameworks such as AutoGen and ROSA rely on structured prompting to enable agentic collaboration in text-based or symbolic domains, they lack embodiment and do not operate in physically grounded environments. The effectiveness of our method benefits from realistic simulation platforms such as Habitat 3.0 \cite{puig2023habitat30cohabitathumans} and VirtualHome \cite{puig2018virtualhome}, which serve as high-fidelity testbeds for embodied AI performance assessment.


Our system is distinguished from previous research by developing and  evaluating   structured prompting techniques combined with LLM configurations.  We demonstrate that this improves agent cooperation in collaborative planning and communication within shared environments.

\section{System Architecture}
We build upon CoELA, which has a modular architecture (see \cite{umass2023coela} for details), where each agent operates independently and is composed of five interconnected modules: Perception, Memory, Planning, Communication, and Execution. 

We focused   on the Planning and Communication modules, which  use LLMs. In the following sections, we explore  different  prompting strategies to enhance task coordination and inter-agent dialogue.

\begin{itemize}
    \item \textbf{Communication Module - } The Communication Module facilitates information exchange between agents, enabling collaborative behaviour in a decentralised setting. It manages the generation of messages related to task status, environment updates, and coordination needs. This module allows agents to share knowledge, request assistance, negotiate roles, and synchronise actions, making it essential for coherent multi-agent interaction and teamwork. 
    \item \textbf{Planning Module - } The Planning Module is responsible for generating action plans that enable agents to achieve specific goals within the environment. It interprets the current world state, objectives, and agent capabilities to produce a structured sequence of high-level decisions. These plans guide the agent’s behaviour over time and provide a foundation for task execution. The module ensures logical consistency, goal alignment, and adaptability to dynamic environmental changes.
\end{itemize}

\section{Ollama Integration}
To support flexible experimentation with different Large Language Models, we integrated Ollama into our system. Ollama \cite{ollama} provides a lightweight, local runtime environment for deploying a wide range of LLMs, including both open-source and quantised models. This integration simplifies the process of switching between models like LLaMA, Mistral, or DeepSeek, allowing us to test various architectures without modifying the core agent pipeline. Additionally, Ollama’s support for quantised models significantly reduces memory overhead and computational cost, enabling efficient execution on resource-constrained systems

\section{Prompt Testing and Optimisation}
The main goal of our project was to develop and test different prompting methods and their combinations to find out which method works best for most of the LLMs and optimises agent collaboration. There are 3 modules in the CoELA setup where we can change prompts: Planning, Communication, and Action. 

In order to conduct   prompt testing,  we utilised four large language models: Llama 3.1 (8B), DeepSeek-v1 (8B), Mistral (7B), and Gemma 3 (4B). These models were selected to balance performance and hardware efficiency. Model sizes were intentionally capped below 8B parameters to ensure that the entire system could run efficiently on a single GPU setup.  Llama 3.1 and DeepSeek are among the most powerful open-source releases, offering state-of-the-art capabilities in reasoning, which makes them particularly well suited for dynamic multi-agent environments. At the same time, using lightweight yet modern models like Gemma 3 allows us to explore whether the number of parameters directly correlates with performance in tasks like planning and coordination. This provides valuable insight into how smaller, more efficient models can contribute meaningfully within collaborative agent setups. All prompts, evaluation metrics, and implementation details discussed below are available on GitHub.

\subsection{Planning Prompt}
The Planning prompt guides the LLM to select the best action for goal completion. CoELA includes a Baseline Planning Prompt that provides the goal, progress, dialogue history, action history, and available actions to the LLM. We tested two modifications:
\begin{itemize}
    \item \textbf{Explicit Instruction --Improved Baseline:} Adds explicit instructions to encourage more goal-focused decisions. For example, instead of asking,{\it ``please help me choose the best available action''} the prompt now says, {\it ``You should select the most efficient action based on the goal, progress, and available options. Ensure your choice contributes directly to goal completion."}
    \item \textbf{Structured/Forced Reasoning:} Extends the prompt with step-by-step questions \footnote{https://github.com/Himajacob/Co-LLM-Agents?tab=readme-ov-file\#a3-structured-base-prompt---forced-reasoning} to guide the LLM toward better decisions by explicitly prompting it to analyse the goal, current state, and available actions.
\end{itemize}

\subsection{Communication Prompt}
The Communication prompt in CoELA enables inter-agent collaboration by generating goal-orientated messages. The Baseline Communication Prompt includes variables like goal, progress, dialogue history, and action history, and ends with a note requesting a brief, accurate message.
\begin{itemize}
    \item \textbf{CPrompt 1 --Instruction Removal:} Removes verbose output by explicitly instructing the LLM to avoid explanations and formatting (e.g., {\it ``Respond as if speaking directly to \$OPPO\_NAME\$"}).
    \item \textbf{CPrompt 2 -- One-Shot:} Adds a single example dialogue \footnote{https://github.com/Himajacob/Co-LLM-Agents?tab=readme-ov-file\#b3-cprompt2---one-shot} to set message tone and structure.
    \item \textbf{CPrompt 3 -- Multi-Shot:} Extends CPrompt 2 with multiple dialogue turns for richer context.
    \item \textbf{CPrompt 4 = CPrompt 1 + CPrompt 2:} Merges the instruction-removal from CPrompt 1 with the one-shot example from CPrompt 2 to generate concise, direct messages that follow a clear format without including unnecessary reasoning or explanations.
\end{itemize}

\subsection{Action Prompt}
The Action prompt is used  to fine-tune how the LLM selects the next best action during execution. It ensures that the model's output aligns with what the system expects, helping to avoid vague or incorrect responses.\begin{itemize}
    \item \textbf{Baseline Prompt:} Simply asks the LLM to generate the next action from the set of given actions.
    \item \textbf{Action Prompt 1 - One Shot:} Improves reliability by adding explicit instructions and an example to help the LLM understand the desired output format. This helps to reduce fuzzy matches and improve efficiency.
\end{itemize}

\section{TTS and Chat GUI Integration}
We developed an interface for the CoELA system which  gives a synchronised visual and auditory interface for displaying real-time dialogue between agents. The purpose of this module is to present  conversations between the agents (Alice and Bob) by reading out a live-updating or already existing conversation log using TTS, and presenting it in a Chat GUI. This aids understanding of system behaviour and supports development. The speech output is currently optimised for clarity and naturalness using the accent and speed features provided along with {\tt gTTS} and {\tt pyttsx3}, with future enhancements aimed at incorporating true emotional expressiveness to improve user interaction.
\begin{figure}[H]
    \centering
    \includegraphics[width=0.6\textwidth]{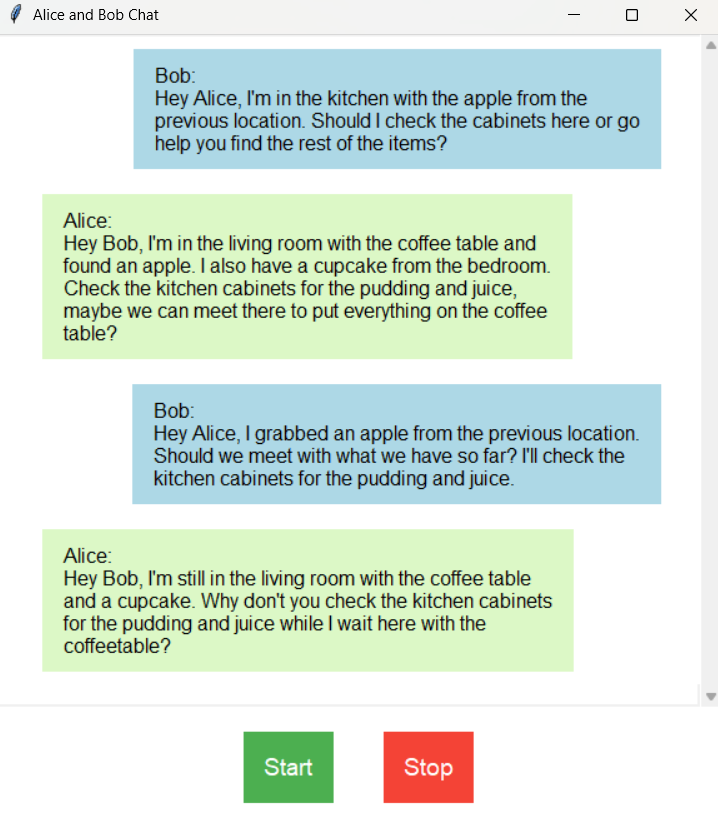}
    \caption{Chat visualisation TTS+ GUI - Alice and Bob Chat}
    \label{fig:Chat GUI Alice and Bob Chat}
    \vspace{-2em}
\end{figure}
The system reads structured dialogue logs, cleans special characters and displays the conversation in a chat-style GUI using {\tt tkinter}. It uses {\tt pyttsx3} for text-to-speech and {\tt pygame} for audio playback, saving each utterance as a temporary .wav file for non-blocking playback. Alice and Bob are assigned natural-sounding female and male voices to enhance clarity and aid understanding.

\section{Evaluation and Results}
We evaluated the  updated CoELA system quantitatively and qualitatively using metrics  such as step count and turn count, and information from the conversation logs. Every prompt combination runs in collaboration mode using  4 different LLMs (Llama3.1:8b, DeepSeek r1:8b, Mistral:7b, Gemma3:4b).
The evaluations  ran over 5 tasks with two variations per task, which together make 10 episodes.   For example a task could be to clear up items and 2 variations could be a)     2 plates and 3 forks, and b)   1 plate and 4 forks.

\vspace{-2em}
\begin{figure}[H]
    \centering
    \includegraphics[width=\textwidth]{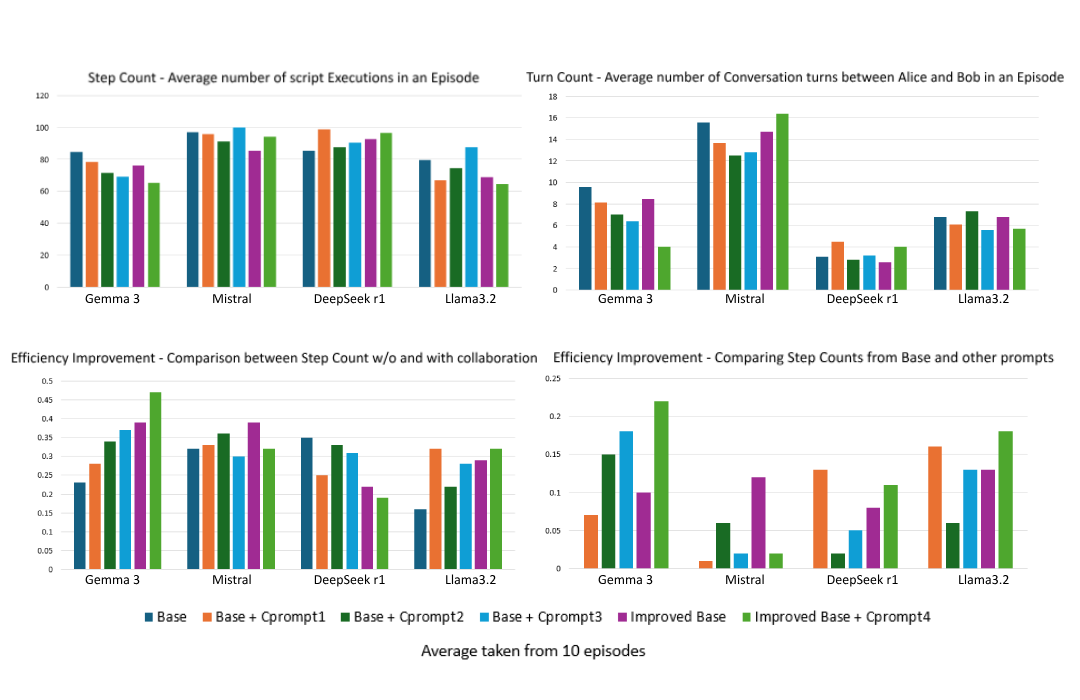}
    \vspace{-2em}
    \caption{
    Consolidated Results from Test Runs (a) Step Count,  (b) Turn Count, (c) Efficiency Improvement 1, (d) Efficiency Improvement 2.}
    \label{fig:Test Runs}
    \vspace{-2em}
\end{figure}

\begin{table}[H]
\centering
\begin{adjustbox}{max width=0.7\textwidth}
\begin{tabular}{|c| c |c |c |c|} 
 \hline
 Prompt & Gemma 3 & Mistral & DeepSeek r1 & Llama 3.1 \\ [0.5ex] 
 \hline\hline
 Base & 84.7 &	97.1 &	85.4 &	79.7 \\
 \hline
 Base + Cprompt1 &	78.3 &	95.9 &	99 &	66.7 \\
 \hline
 Base + Cprompt2 &	71.6 &	91.1 &	87.7 &	74.5 \\
 \hline
 Base + Cprompt3 &	69.3 &	100.0 &	90.28 &	68.8 \\
 \hline
 Base + Cprompt4 &	87.7 &	94.5 &	89.7 &	65.4 \\
 \hline
 Improved Base &	76.2 &	85.3 &	92.8 &	68.7 \\
 \hline
 Improved Base + Cprompt1 &	74.5 &	98.1 &	89.0 &	69.7 \\
 \hline
 Improved Base + Cprompt4 &	\cellcolor[HTML]{50C878} 65.5 &	94.4 &	96.5 &	\cellcolor[HTML]{50C878} 64.6 \\
 \hline
 Base + Cprompt4 + action one shot &	72.2 &	102.75 &	93.8 &	66.3 \\
 \hline
\end{tabular}
\end{adjustbox}
\vspace{1em}
\caption{Step Count Comparison (lower is better)}
\vspace{-1em}
\label{tab:step-count}
\end{table}
The first  metric, {\it step count}, is the number of script executions for successfully completing a task. We ran the system for all the 10 episodes and took the average step count for each LLM and prompt combination as shown in Table \ref{tab:step-count}. Our goal is to reduce the step count so that tasks are done faster. Upon testing with different prompt combinations, we were able to find that step count reduction is better ensured by the Improved Base and Cprompt4 combination especially with Llama3.1 and Gemma3, as displayed in Figure \ref{fig:Test Runs} (a) and Table \ref{tab:step-count}. For Gemma 3, the step count decreased from 84.0 to 65.5, demonstrating a significant improvement in execution efficiency. Although Llama 3.1 already exhibited strong baseline performance with a step count of 79.0, the optimized prompt combination further reduced it to 64.6. These results suggest that prompt engineering plays a critical role in enhancing agent collaboration performance, even when using lightweight or already well-performing models.

\begin{table}[H]
\centering
\begin{adjustbox}{max width=0.7\textwidth}
\begin{tabular}{|c|| c| c| c| c|} 
 \hline
 Prompt & Gemma 3 & Mistral & DeepSeek r1 & Llama 3.1 \\ [0.5ex] 
 \hline\hline
 Base & 9.6 & 15.6 & 3.1 & 6.8 \\
 \hline
 Base + Cprompt1 & 8.1 & 13.7 & 4.5 & 6.1 \\
 \hline
 Base + Cprompt2 & 7 & 12.5 & 2.8 & 7.3 \\
 \hline
 Base + Cprompt3 & 6.4 & 12.8 & 3.2 & 5.6 \\
 \hline
 Base + Cprompt4 & 8.4 & 14.5 & 3.5 & 6 \\
 \hline
 Improved Base &	8.5 & 14.7 & 2.6 & 6.8 \\
 \hline
 Improved Base + Cprompt1 & 13 & 19.4 & 3.9 & 5 \\
 \hline
 Improved Base + Cprompt4 &	\cellcolor[HTML]{50C878} 4 & 16.4 & \cellcolor[HTML]{50C878} 4 & \cellcolor[HTML]{50C878} 5.7 \\
 \hline
 Base + Cprompt4 + action one shot &	7.8 & 14.5 & 2.7 & 5.4 \\
 \hline
\end{tabular}
\end{adjustbox}
\vspace{1em}
\caption{Turn Count Comparison (lower is better)}
\label{tab:turn-count}
\vspace{-2em}
\end{table} 
The second metric, {\it turn count}, is the number of dialogue turns between the agents Alice and Bob during the collaborative execution of the task. Collaboration is more successful when understanding is better:%
 thus, if the agents are able to understand precisely what they have to do in  fewer turns, that paves the way for efficient execution. With this idea, we developed the prompts and tested with different LLMs, and found that turn counts are also reduced significantly when using the Improved base and Cprompt4 combination, as shown in Figure \ref{fig:Test Runs} (b) and Table \ref{tab:turn-count}.

\begin{table}[H]
\centering
\begin{adjustbox}{max width=0.7\textwidth}
\begin{tabular}{|c|| c| c |c |c|} 
 \hline
 Prompt & Gemma 3 & Mistral & DeepSeek r1 & Llama 3.1 \\ [0.5ex] 
 \hline\hline
 Base & 0.23 & 0.32 & 0.35 & 0.16 \\
 \hline
 Base + Cprompt1 & 0.28 & 0.33 & 0.25 & 0.32 \\
 \hline
 Base + Cprompt2 & 0.34 & 0.36 & 0.33 & 0.22 \\
 \hline
 Base + Cprompt3 & 0.37 & 0.30 & 0.31 & 0.28 \\
 \hline
 Base + Cprompt4 & 0.20 & 0.33 & 0.32 & 0.31 \\
 \hline
 Improved Base & 0.39 & 0.39 & 0.22 & 0.29 \\
 \hline
 Improved Base + Cprompt1 & 0.40 & 0.30 & 0.32 & 0.27 \\
 \hline
 Improved Base + Cprompt4 &	\cellcolor[HTML]{50C878} 0.47 & 0.32 & 0.19 & \cellcolor[HTML]{50C878} 0.32 \\
 \hline
 Base + Cprompt4 + action one shot &	0.34 & 0.28 & 0.28 & 0.30 \\
 \hline
\end{tabular}
\end{adjustbox}
\vspace{1em}
\caption{Efficiency Improvement due to collaboration}
\vspace{-2em}
\label{tab:efficiency1}
\end{table}

One key observation was that the base prompt often led to verbose outputs, with LLMs including internal ``thinking" alongside actual messages, which reduced clarity. In contrast, the revised prompts encouraged more concise, action-focused communication. However, some models like DeepSeek r1 continued to generate reasoning even with restricted prompts, suggesting that it is difficult to fully bypass an LLM's inherent behaviour through prompting alone. Overall, prompt refinement significantly improved both communication clarity and task execution efficiency.

\begin{table}[H]
\centering
\begin{adjustbox}{max width=0.7\textwidth}
\begin{tabular}{|c| c |c |c| c|} 
 \hline
 Prompt & Gemma 3 & Mistral & DeepSeek r1 & Llama 3.1 \\ [0.5ex] 
 \hline\hline
 Base + Cprompt1 & 0.07 & 0.01 & 0.13 & 0.16 \\
 \hline
 Base + Cprompt2 & 0.15 & 0.06 & 0.02 & 0.06 \\
 \hline
 Base + Cprompt3 & 0.18 & 0.02 & 0.05 & 0.13 \\
 \hline
 Base + Cprompt4 & 0.03 & 0.02 & 0.04 & 0.17 \\
 \hline
 Improved Base & 0.10 & 0.12 & 0.08 & 0.13 \\
 \hline
 Improved Base + Cprompt1 & 0.12 & 0.01 & 0.04 & 0.12 \\
 \hline
 Improved Base + Cprompt4 &	\cellcolor[HTML]{50C878} 0.22 & 0.02 & \cellcolor[HTML]{50C878} 0.11 & \cellcolor[HTML]{50C878} 0.18 \\
 \hline
 Base + Cprompt4 + action one shot &	0.14 & 0.05 & 0.08 & 0.16 \\
 \hline
\end{tabular}
\end{adjustbox}
\vspace{1em}
\caption{Efficiency Improvement due to new prompts}
\vspace{-2em}
\label{tab:efficiency2}
\end{table}

The efficiency improvement of the system is measured based on the step count data gathered from the experiments. It is measured in two ways, where:
\begin{itemize}
    \item \textbf{Efficiency Improvement 1} -- The step count values collected from collaboration runs are compared to those of individual agent runs. This indicates how collaboration impacts the execution of a task. From the data shared in table \ref{tab:efficiency1} and the graph in Figure \ref{fig:Test Runs}(c), we can see that collaboration always positively affects the system and improves efficiency. Here, the best improvement is shown by the Improved base and Cprompt4 combination.  In particular, Gemma 3 demonstrated a substantial improvement -- a 47\% reduction in step count when shifting from single-agent to collaborative execution under the optimised prompting strategy. This clearly reinforces the value of effective inter-agent communication and prompt engineering in boosting collaborative performance. 
    \item \textbf{Efficiency Improvement 2} --  compares step counts from new prompting strategies against the base prompt. While the Improved Base + Cprompt4 combination showed notable gains for Gemma 3 (22\%) and Llama 3.1 (18\%), the same improvement was not consistent across all models. 
    \end{itemize}
    
    \subsection{Discussion}
\begin{figure}[H]
    \centering
    \includegraphics[width=\textwidth]{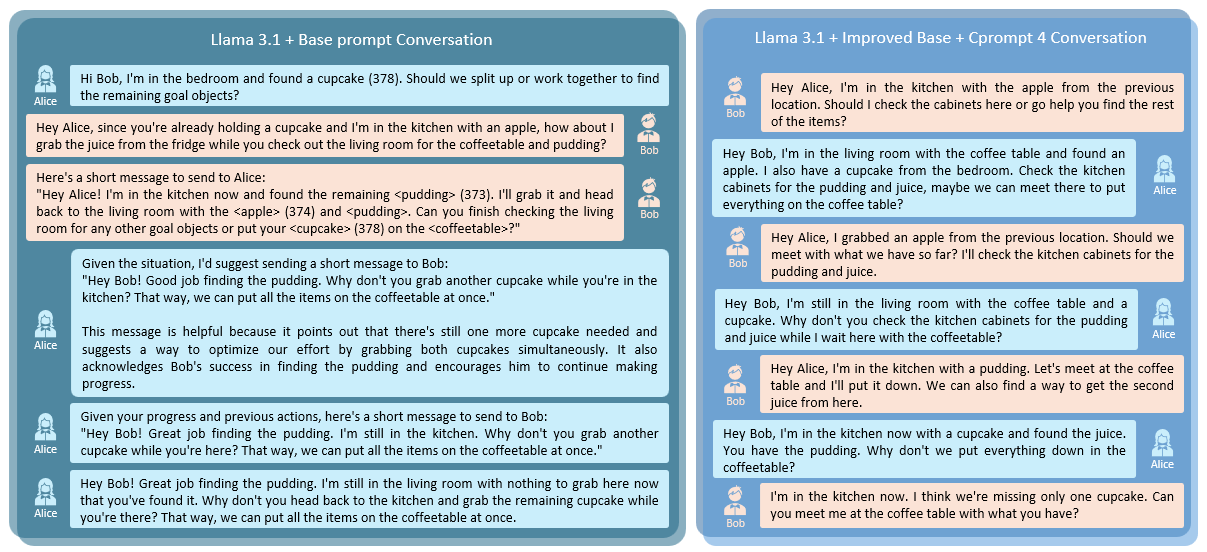}
    \vspace{-2em}
    \caption{Conversation Comparison  (left: baseline, right: improved)}
    \vspace{-2em}
    \label{fig:Conversation Comparision}
\end{figure}

    The best performing configuration  in our experiments was the combination of the Improved Baseline prompt with Cprompt4, which consistently reduced the step count for both Llama 3.1 and Gemma 3.
    See figure \ref{fig:Conversation Comparision} for an illustration of the differences in collaborative behaviour. While this combination proved effective across models, one key insight from our testing was that prompting efficiency is highly model dependent. For instance, DeepSeek, even when provided with few-shot examples in the communication prompt, tended to include its internal reasoning or ``thinking process" in its responses. This behaviour persisted despite modifications in the prompt structure, indicating that DeepSeek's response style is less sensitive to prompt engineering than other models.

    Another interesting observation was that Gemma 3, despite being a smaller model (4B parameters), delivered performance comparable to Llama 3.1 (8B parameters). This suggests that newer models like Gemma may incorporate architectural or training improvements that allow them to match or even outperform larger predecessors in certain collaborative scenarios. To statistically validate the observed performance improvements, we conducted a t-test comparing the Base prompt against the Improved Base + Cprompt4 configuration. The results showed a t-score of 1.70 (p = 0.62) for Gemma 3, and a t-score of 2.47 (p = 0.018) for Llama 3.1. While Gemma's result suggests only a moderate and not statistically significant improvement, Llama’s higher t-score and low p-value indicate a statistically significant effect from the prompt refinement. These findings reinforce the earlier observation that Llama 3.1 benefits more consistently from prompt optimization, whereas Gemma 3, despite its smaller size, still shows substantial practical gains due to its more recent architectural improvements.
%

\vspace{-2em}

\section{Conclusion and Future Work}
\vspace{-0.25em}
In this work, we enhanced the CoELA collaborative embodied AI system by developing improved prompting strategies alongside a real-time dialogue interface. The combination of Explicit Instruction (Improved Base) and CPrompt 4 significantly reduced verbosity, redundant actions, and fuzzy (i.e., ambiguous or improperly formatted) LLM outputs, leading to more efficient inter-agent communication and task execution. Along with these improvements, to aid development, we implemented a TTS-ChatGUI module that shows and audibly pronounces the conversation in real time. This interface makes differences in clarity, turn-taking, and task alignment visibly and audibly apparent, and so aids system development. Together, our results highlight the value of integrating prompt engineering with interactive, multimodal agent interfaces.

In future developments we plan to extend our evaluation beyond agent-agent collaboration to include human-agent interaction using similar prompting strategies, as well as to ad-hoc team formations of $n \geq 3$ agents (e.g.\ 2 robots assisting 1 human). We are particularly interested in examining how well agents respond to natural human communication and to dynamic team formation. Understanding these phenomena will help us improve collaborative agent behaviour in real-world, mixed-team settings.

A demonstration video is available at \href{https://youtu.be/MtP8BH7kYsE}{[ https://youtu.be/MtP8BH7kYsE ]} and the code is available at 
\href{https://github.com/Himajacob/Co-LLM-Agents}{[https://github.com/Himajacob/Co-LLM-Agents]}.

\bibliography{anthology,custom}

\begin{thebibliography}{10}

\bibitem{agashe2024agents}
Saaket Agashe, Jiuzhou Han, Shuyu Gan, Jiachen Yang, Ang Li, and Xin~Eric Wang.
\newblock Agent s: An open agentic framework that uses computers like a human.
\newblock {\em arXiv preprint arXiv:2410.08164}, 2024.

\bibitem{deepseekai2025deepseekr1incentivizingreasoningcapability}
DeepSeek-AI and Daya~Guo et~al.
\newblock {DeepSeek-R1: Incentivizing Reasoning Capability in LLMs via Reinforcement Learning}, 2025.

\bibitem{deitke2022retrospectives}
Matthias Deitke, Dhruv Batra, Yonatan Bisk, Tommaso Campari, Angel~X. Chang, Devendra~Singh Chaplot, ..., and Jiajun Wu.
\newblock {Retrospectives on the Embodied AI Workshop}.
\newblock {\em arXiv preprint arXiv:2210.06849}, 2022.

\bibitem{grattafiori2024llama3herdmodels}
Aaron~Grattafiori et~al.
\newblock The llama 3 herd of models, 2024.

\bibitem{jiang2023mistral7b}
Albert Q.~Jiang et~al.
\newblock Mistral 7b, 2023.

\bibitem{hassouna2024llm}
Amine~Ben Hassouna, Hana Chaari, and Ines Belhaj.
\newblock Llm-agent-umf: Llm-based agent unified modeling framework for seamless integration of multi active/passive core-agents.
\newblock {\em arXiv preprint arXiv:2409.11393}, 2024.

\bibitem{ollama}
Ollama Inc.
\newblock Ollama: Get up and running with large language models, 2025.
\newblock Accessed: 2025-04-04.

\bibitem{li2023camelcommunicativeagentsmind}
Guohao Li, Hasan Abed Al~Kader Hammoud, Hani Itani, Dmitrii Khizbullin, and Bernard Ghanem.
\newblock {CAMEL: Communicative Agents for "Mind" Exploration of Large Language Model Society}, 2023.

\bibitem{mandi2023rocodialecticmultirobotcollaboration}
Zhao Mandi, Shreeya Jain, and Shuran Song.
\newblock {RoCo: Dialectic Multi-Robot Collaboration with Large Language Models}, 2023.

\bibitem{autogen2023}
{Microsoft AutoGen Team}.
\newblock {AutoGen: Enabling Next-Gen LLM Applications via Multi-Agent Conversation Framework}.
\newblock \url{https://microsoft.github.io/autogen/0.2/}, 2023.

\bibitem{puig2018virtualhome}
Xavier Puig and et~al.
\newblock {VirtualHome: Simulating Household Activities via Programs}.
\newblock \url{http://virtual-home.org/}, 2018.

\bibitem{puig2023habitat30cohabitathumans}
Xavier Puig, Eric Undersander, Andrew Szot, Mikael~Dallaire Cote, Tsung-Yen Yang, Ruslan Partsey, Ruta Desai, Alexander~William Clegg, Michal Hlavac, So~Yeon Min, Vladimír Vondruš, Theophile Gervet, Vincent-Pierre Berges, John~M. Turner, Oleksandr Maksymets, Zsolt Kira, Mrinal Kalakrishnan, Jitendra Malik, Devendra~Singh Chaplot, Unnat Jain, Dhruv Batra, Akshara Rai, and Roozbeh Mottaghi.
\newblock {Habitat 3.0: A Co-Habitat for Humans, Avatars and Robots}, 2023.

\bibitem{royce2025enablingnovelmissionoperations}
Rob Royce, Marcel Kaufmann, Jonathan Becktor, Sangwoo Moon, Kalind Carpenter, Kai Pak, Amanda Towler, Rohan Thakker, and Shehryar Khattak.
\newblock {Enabling Novel Mission Operations and Interactions with ROSA: The Robot Operating System Agent}, 2025.

\bibitem{shinn2023reflexionlanguageagentsverbal}
Noah Shinn, Federico Cassano, Edward Berman, Ashwin Gopinath, Karthik Narasimhan, and Shunyu Yao.
\newblock {Reflexion: Language Agents with Verbal Reinforcement Learning}, 2023.

\bibitem{gemmateam2025gemma3technicalreport}
Gemma Team and Aishwarya~Kamath et~al.
\newblock Gemma 3 technical report, 2025.

\bibitem{umass2023coela}
{UMass Embodied AGI Lab}.
\newblock {CoELA GitHub Repository}.
\newblock \url{https://github.com/UMass-Embodied-AGI/CoELA}, 2023.

\bibitem{wang2023voyageropenendedembodiedagent}
Guanzhi Wang, Yuqi Xie, Yunfan Jiang, Ajay Mandlekar, Chaowei Xiao, Yuke Zhu, Linxi Fan, and Anima Anandkumar.
\newblock {Voyager: An Open-Ended Embodied Agent with Large Language Models}, 2023.

\bibitem{wang2024describeexplainplanselect}
Zihao Wang, Shaofei Cai, Guanzhou Chen, Anji Liu, Xiaojian Ma, and Yitao Liang.
\newblock {Describe, Explain, Plan and Select: Interactive Planning with Large Language Models Enables Open-World Multi-Task Agents}, 2024.

\bibitem{yao2023treethoughtsdeliberateproblem}
Shunyu Yao, Dian Yu, Jeffrey Zhao, Izhak Shafran, Thomas~L. Griffiths, Yuan Cao, and Karthik Narasimhan.
\newblock Tree of thoughts: Deliberate problem solving with large language models, 2023.

\bibitem{yao2023reactsynergizingreasoningacting}
Shunyu Yao, Jeffrey Zhao, Dian Yu, Nan Du, Izhak Shafran, Karthik Narasimhan, and Yuan Cao.
\newblock {ReAct: Synergizing Reasoning and Acting in Language Models}, 2023.

\bibitem{zhang2024buildingcooperativeembodiedagents}
Hongxin Zhang, Weihua Du, Jiaming Shan, Qinhong Zhou, Yilun Du, Joshua~B. Tenenbaum, Tianmin Shu, and Chuang Gan.
\newblock Building cooperative embodied agents modularly with large language models, 2024.

\end{thebibliography}
\bibliographystyle{plain}

\end{document}